\documentclass[conference]{IEEEtran}
\usepackage{subfigure}
\usepackage{graphicx}
\usepackage{float}
\ifCLASSINFOpdf
\else
\fi
\begin{document}
%
\title{A Parallel Way to Select the Parameters of SVM Based on the Ant Optimization Algorithm}

\author{\IEEEauthorblockN{Chao Zhang}
\IEEEauthorblockA{Software Engineering\\Software College\\SiChuan University,China\\
Email: zhch040200@gmail.com}
\and
\IEEEauthorblockN{Hong-Cen Mei}
\IEEEauthorblockA{Software Engineering\\Software College\\SiChuan University,China\\
Email: hongcenmei@yeah.net}
\and
\IEEEauthorblockN{Hao Yang}
\IEEEauthorblockA{Software Engineering\\Software College\\SiChuan University,China\\
Email: chongqingyanghao@yeah.net
}}


%


\maketitle

\begin{abstract}
A large number of experimental data shows that Support Vector Machine (SVM) algorithm has obvious a large advantages in text classification, handwriting recognition, image classification, bioinformatics, and some other fields. To some degree, the optimization of SVM depends on its kernel function and Slack variable, the determinant of which is its parameters  $\delta$ and c in the classification function. That is to say, to optimize the SVM algorithm, the optimization of the two parameters play a huge role. Ant Colony Optimization (ACO) is optimization algorithm which simulate ants to find the optimal path. In the available literature, we mix the ACO algorithm and Parallel algorithm together to find a well parameters.
\end{abstract}


%
\IEEEpeerreviewmaketitle
\section{Support Vector Classification and Parameters}
SVM is based on the principle of structural risk minimization,using limited training samples to obtain the higher generalization ability of decision function. Suppose a sample set $(x_i,y_i)$ , where $i=1,2_{...}N$ means the number of training samples, $x{\in}R$ means the sample characteristics, $y\in\{+1,-1\}$ means the sample classification.\\
SVM Classification function:
\begin{center}
$y=\omega x+b$
($\omega$ means weight vector,b means setover)
\end{center}
Functional margin :
$$\gamma_{1} =y({¦Ø}x+b)=yf(x)$$
\begin{center}
functional margin is the minimum margin from hyperplane $(\omega,b)$ to $T(x_i,y_i)$\\
\end{center}
Geometrical margin:
$$\gamma_{2}=\frac{y({\omega}x+b)}{\parallel{\omega}\parallel}=\frac{{\mid}f(x){\mid}}{\parallel{\omega}\parallel}$$
When classifying a data point, the larger the margin is, the more credible the classification is. So to improve the credibility is to maximize the margin.  The  $\omega$ and $b$ can be proportional scaled through the functional margin, thus the value of $y={\omega}x+b$ can be any large. Such that it is not appropriate to maximize a value. But in the geometrical margin, when scaling the $\omega$ and b, the value of $\gamma_2$ can't be change. So it is appropriate to maximize a value.\\
Slack variable:\\
$\varepsilon_i>0$ Using to allow the data to deviate from the hyperplane to a certain extent.\\
Radial Basis Function kernel:
$$k{\langle}x_1,x_2\rangle=e^{-{\frac{{{\parallel}x_1-x_2\parallel}^2}{2\sigma^2}}}$$
Maximum margin classifier:
$$MAX \gamma_2 \quad s.t  \gamma_{1_{i}}\geq{\gamma_{1}-\varepsilon_i},i=1,2,3_{...}n$$
Let: $\gamma_{1}=1$,so
$$MAX \frac{1}{\parallel{\omega}\parallel}+C{\sum_{i=1}^{n}}{\varepsilon}_{i} s.t \gamma_{1_{i}}\geq{\gamma_{1}-\varepsilon_i},i=1,2,3_{...}n$$
The constraints is associated with objective function through the Lagrange function:
$$L(\omega,b,\alpha)=\frac{1}{2}{\parallel{\omega}\parallel}^2-{\sum_{i=1}^{n}}\alpha_i(y_i(({\omega}x+b)-1)+C{\sum_{i=1}^{n}}{\varepsilon}_{i}$$
Let:
$$\Theta(\omega)= MAX  L(\omega,b,\alpha)$$
When all the constraints is contented, the $\Theta(\omega)=\frac{1}{2}\parallel{\omega}\parallel^2$ is the value that we first want to minimized.So objective function:
$$MIN\Theta(\omega)=MINMAX L(\omega,b,\alpha)$$
Dual function:
$$MIN\Theta(\omega)=MAXMIN L(\omega,b,\alpha)$$
To solve the problem, requiring:
$$\frac{{\partial}L}{{\partial}\omega}=0\Rightarrow\omega=\sum_{i=1}^{n}{\alpha}_ix_iy_i$$
$$\frac{{\partial}L}{{\partial}b}=0\Rightarrow\sum_{i=1}^{n}{\alpha}_iy_i=0$$
$$\frac{{\partial}L}{{\partial}{\varepsilon}_i}=0\quad means \quad C-{\alpha}_i-{\gamma}_i=0,i=1,2_{...}n$$
$$MAXMINL(\omega,b,\alpha)=\frac{1}{2}{\parallel{\omega}\parallel}^2-{\sum_{i=1}^{n}}\alpha_i(y_i(({\omega}x+b)-1)$$
$$=MAX\sum_{i=1}^{n}{\alpha}_i-{\frac{1}{2}}\sum_{i,j=1}^{n}{\alpha}_{i}{\alpha}_{j}y_{i}y_{j}k{\langle}x_1,x\rangle$$
The minimized duel problem:
$$MAX\sum_{i=1}^{n}{\alpha}_i-{\frac{1}{2}}\sum_{i,j=1}^{n}{\alpha}_{i}{\alpha}_{j}y_{i}y_{j}k{\langle}x_1,x\rangle$$
$$s.t0\le{\alpha}_{i}\le{C},i=1,2,3_{...}n;\sum_{i=1}^{n}{\alpha}_iy_i=0$$
The final classification function:
$$y={\omega}x+b$$
$$\sum_{i=1}^{n}{\alpha}_{i}y_{i}k{\langle}x_1,x\rangle+b$$
\section{Modified Ant Colony Optimization Algorithm}
Different from traditional problem of TCP, in this algorithm,the coordinate is used to represented the node. In a two-dimensional rectangular coordinate system, the significance of X is defined as the significant digit of parameters C and $\delta$, Y is varied from 0 to 10.\cite{article2} The significant digit of C are assumed to be the five, and the highest level of C is hundred's place. Similarly, assume that the significant digit of  $\delta$  are also assumed to be the five, and the highest level of $\delta$  is The Unit.\\
To realize the ant colony optimization, we follow the steps below:
\begin{enumerate}
\item Suppose there are $m$ ants. Each ant $k(k=1{\sim}m)$has a one-dimensional array $Path(k)$ which has n elements(n is the total significant digit of C and $\delta$), it¡¯s used to store the vertical coordinates  $y$ of each point which the ant $k$ visited.
\item Set the loop time N=0 to $N_{max}$. Initialize the each point¡¯s pheromone concentration $\tau(x,y)=\tau_0 (x=0\sim n,y=0\sim9)$.Set $x=1$.
\item Set $y=1$ to $n$.Calculate the deflection probability $P_k$ of each ant move to the node vertical line $L_x$.Then select the next point via roulette wheel and store the point¡¯s vertical coordinates y to $Path_k$
$$P_k(x,y)=\frac{\tau^{\alpha}(x,y)\eta^\beta(x,y)}{\sum_{j=0}^{9}\tau^{\alpha}(x,y)\eta^\beta(x,y)}$$
$$\tau(x,y)=\rho\tau(x,y)+\Delta\tau(x,y)$$
$$\Delta\tau(x,y)=\sum_{k=1}^{m}\Delta\tau_k(x,y)$$
$$\Delta\tau(x,y)=\left\{\begin{array}{ll}\frac{Q}{1-Acc_k}&\textrm{,if ant k visited the point(x,y)}\\0& \textrm{,others}\end{array} \right.$$
\begin{center}
The $Acc_k$ mean the ant k s accuracy of  cross-validation\\
$P_k$ means the deflection probability point$(x-1,y)$ to $(x,y)$\\
Q is a constant\\
W:the weight coefficient;\\
Set:the minimum acceptable accuracy;\\
\end{center}
\item Let :$x=x+1$,if $x\le{n}$,turn to step 3;else turn to step 5.
\item Record this motion path $Path_k$ calculate the mapping data $(c,\delta)$.
\item Make the training samples evenly divided into k mutually exclusive subsets of $S_1,S_2,...,S_k;$
\item Calculate the cross-validation accuracy:
\begin{enumerate}
\item Initialize i=1;
\item Make the $S_i$ subset reserved for test sets, and set the rest as the training set, training SVM;
\item Calculate the $i^{th}$ subset¡¯s Sample Classification Accuracy $Acc_i$,set $i=i+1$,When $ i <k +1$, repeat the step b);
\item Calculate the mean of k Sample Classification Accuracy $\hat{Acc}$ :
\end{enumerate}
$$Acc=\frac{Right}{Error+Right}$$
$$\widehat{Acc}=\frac{\sum_{i=1}^{k}Acc_i}{k}$$
\begin{center}
Right: Correctly classified (+1) number of samples\\
Error:Misclassification (-1) number of samples\\
$\hat{Q}$ :the mean of  Sample Classification Accuracy
\end{center}
\item Update the pheromone concentration $\tau(x,y)$ at each points by $\hat{Acc}$ . Clear $Path(k_i)$;
\item Reset $N=N+1$.When $N\le{N_{max}}$ , and the entire colony has not converged to follow the same path, then turn to step 3; $N\le{N_{max}}$, and the entire colony substantially converged to follow the same path, then the algorithm ends. Take the last update of the path and its mapping data to $(c,\delta)$,that  is the SVM parameters C, $\delta$ final optimization results.
\end{enumerate}
In order to validate the algorithm,we do simulation experiment on matlab R2010b and PC with windows7 64-bit operating system, 4 G memory, core i5 processor. We divide the $wine\_SVM$ data set into a sample set and a test set. Then we will have 90 training datas and 88 test datas. We set ACO parameter m=30, N=500, $\rho$=0.7, Q=100, $\alpha$=1, $\beta$=1. In order to calculate the classification accuracy, we introduced the LIBSVM.\cite{article3}.LIBSVM has gained wide popularity in machine learning and many other areas.\cite{article1}
\begin{figure}[h]
\subfigure[The best accuracy's path] {\includegraphics[height=1.7in,width=1.7in,angle=0]{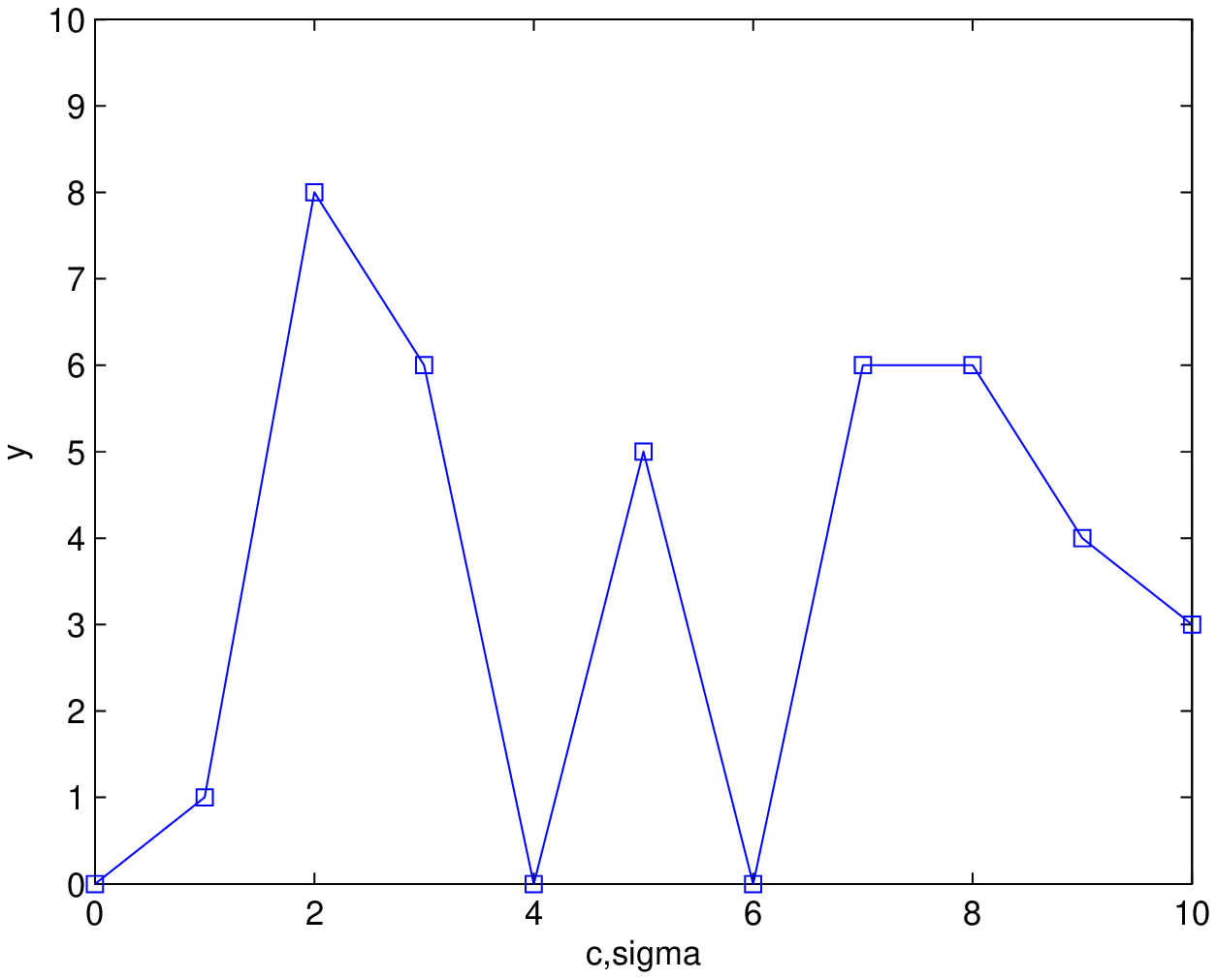}}
\subfigure[All accuracy's paths] {\includegraphics[height=1.7in,width=1.7in,angle=0]{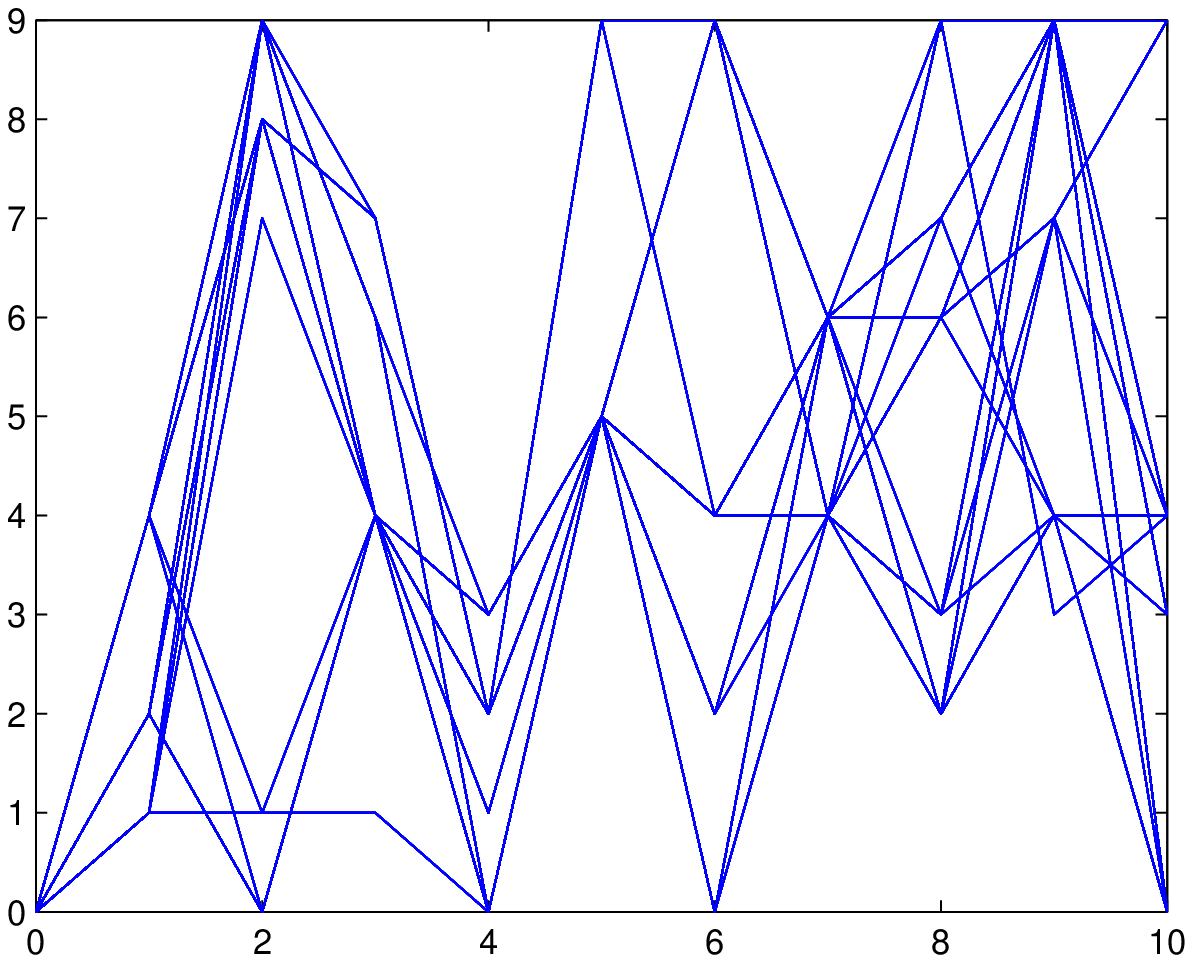}}
\end{figure}
\\The best accuracy is 95.4545\% and the convergence accuracy is 86.3636\% (76/88).
\\The parameter c: 18.605, parameter $\sigma$: 0.6643
\section{Parallel Optimize the Parameters}
The Open Computing Language, is an open specification for heterogeneous computing released by the Khronos Group2 in 2008. It resembles the NVIDIA CUDA3 platform, but can be considered as a superset of the latter, they basically differ in the following points\cite{book}:
\begin{itemize}
\item OpenCL is an open specification that is managed by a set of distinct representatives from industry, software development, academia and so forth.
\item OpenCL is meant to be implemented by any compute device vendor, whether they produce CPUs, GPUs, hybrid processors, or other accelerators such as digital signal processors (DSP) andfield-programmable gate arrays (FPGA).
\item OpenCL is portable across architectures, meaning that a parallel code written in OpenCL is guaranteed to correctly run on every other supporte device.
\end{itemize}
In this ant colony optimization algorithm, we have $m$ ants and we loop it $N\_max$ times. If we give $m$ a large number, such as ten thousand or one hundred thousand, our update of node pheromone concentration will be more accurate and reliable, but we will spend more time. As we all know, every cycle, each ant's access to nodes are unrelated with others,so we can parallel the ant access process. In this article, we use openCL to realize the parallel of ant\cite{book}.
\begin{center}
\begin{tabular}{l}
\\
\hline
OpenCL kernel for the ant-based solution construction\\
\hline
$global_{size}=number_{ants}$;\\
$visited[globale_{size}\times{number_{nodes}}]=\{-1\}$;\\
\\
for ( i=0 to n-1) do\\
$\{$\\
\quad $sum_{prob}=0$;\\
\quad for (j=0 to 9) do\\
\quad $\{$\\
\quad \quad $selection_{prob}[global_{id}\times{number_{nodes}}+j]$\\
\quad \quad \quad $=choice_{info}[i\times{n}+j]$;\\
\quad \quad $Sum_{prob}=sum_{prob}+$\\
\quad \quad \quad $selection_{prob}[global_{id}\times{number_{nodes}}+j];$\\
\quad $\}$\\
\quad j=0;\\
\quad $p=selection_{prob}[global_{id}\times{number_{nodes}}+j];$\\
\quad while  $p<random(0,sum_{prob})$ do \\
\quad $\{$\\
\quad \quad j=j+1;\\
\quad \quad $p=p+selection_{prob}[global_{id}\times{number_{nodes}}+j]$;\\
\quad $\}$\\
\quad $visited[global_{id}\times{number_{nodes}}+i]=j$;\\
$\}$\\
\hline
\\
\end{tabular}
\end{center}
Training samples are divided into subset$\_$number subset on average and each subset has sample$\_$number samples. One subset$ (S_i)$ see as test set and the other subsets $(S_{1\sim{i}},S_{i\sim{k}})$ see as a training set(each subset has c samples), then according to the current parameters $(c,\delta)$training the SVM, calculating error of K-flod cross validation.
\\
\\
\begin{tabular*}{8.8cm}{l}
\hline
OpenCL kernel for sample classification accuracy\\
\hline
$group_{size}=subset\_number$;\\
$local_{size}=sample\_number$;\\
\\
for (i=0 to $group_{size}-1$)\\
$\{$\\
\quad $Right=Error=Q=0;$\\
\quad $for (j=0$ to $local_{size}-1)$\\
\quad $\{$\\
\quad \quad $f(x)=sign(\sum_{i=1}^{n}\alpha_iy_ik\langle{group_{id}},local_{id}\rangle{+b})$;\\
\quad \quad if\quad $f(x)=1$\\
\quad \quad \quad $Right++$;\\
\quad \quad esle\\
\quad \quad \quad $Error++$;\\
\quad $\}$\\
\quad $Q=Q+\frac{Right}{Right+Error}$;\\
$\}$\\
$Q=Q/group_{size}$;\\
\hline
\end{tabular*}
\section{Conclusion}
Through the ant colony optimization algorithm, we can find a satisfactory parameter of SVM, and the convergence accuracy can be guaranteed  more than 85\%. There are also many other ways to optimize parameters, such as Genetic algorithm (GA)\cite{article4}, dynamic encoding algorithm \cite{article5} for handwritten digit recognition, Particle swarm optimization(PSO)\cite{article6}. We also can parallel Ant Colony Optimization,article \cite{article7} introduced a new way which parallel Ant Colony Optimization on  Graphics Processing Units.Article \cite{article8} improving ant colony optimization algorithm for data clustering.
\section*{Acknowledgment}


%

\end{document}